\documentclass[runningheads]{llncs}
\usepackage[T1]{fontenc}
\usepackage{graphicx,verbatim}
\usepackage{lineno}
\usepackage{cite}
\usepackage{amsmath}
\usepackage{marvosym}
\usepackage{hyperref}
\hypersetup{
    colorlinks=true,
    urlcolor=blue,
    linkcolor=blue,
    citecolor=blue
}
\begin{document}
\title{High-Precision Mixed Feature Fusion Network 
 Using Hypergraph Computation for Cervical Abnormal Cell Detection}
 \titlerunning{High-Precision Cervical Abnormal Cell Detection}
%

\author{Jincheng Li\inst{1}, Danyang Dong\inst{1}, Menglin Zheng\inst{1}, Jingbo Zhang\inst{1}, \\Yueqin Hang\inst{1}, Lichi Zhang\inst{2}, and Lili Zhao\textsuperscript{1(\Letter)}}  
\authorrunning{J Li et al.}
\institute{1.School of Artificial Intelligence and Computer Science,\\Nantong University, Nantong 226019, China \\
    \email{ylzh@ntu.edu.cn}\\
    2.School of Biomedical Engineering, Shanghai Jiao Tong University, \\Shanghai, 200030, China}
    
\maketitle              
\begin{abstract}
Automatic detection of abnormal cervical cells from Thin-prep Cytologic Test (TCT) images is a critical component in the development of intelligent computer-aided diagnostic systems. However, existing algorithms typically fail to effectively model the correlations of visual features, while these spatial correlation features actually contain critical diagnostic information. Furthermore, no detection algorithm has the ability to integrate inter-correlation features of cells with intra-discriminative features of cells, lacking a fusion  strategy for the end-to-end detection model. In this work, we propose a hypergraph-based cell detection network that effectively fuses different types of features, combining spatial correlation features and deep discriminative features. Specifically, we use a Multi-level Fusion Sub-network (MLF-SNet) to enhance feature extraction capabilities. Then we introduce a Cross-level Feature Fusion Strategy with Hypergraph Computation module (CLFFS-HC), to integrate mixed features. Finally, we conducted experiments on three publicly available datasets, and the results demonstrate that our method significantly improves the performance of cervical abnormal cell detection. The code is publicly available at \url{https://github.com/ddddoreen/HyperMF2-Cell-Detection}.


\end{abstract}

\section{Introduction}

Cervical cancer is ranked as the fourth most common cancer globally \cite{Siegel2021}. If diagnosed early, it can be effectively treated and cured \cite{Schiffman2007}. Thin-prep Cytologic Test (TCT) is the most widely utilized and effective screening method for detecting cervical abnormal and precancerous lesions \cite{Davey2006}. Despite its widespread clinical application and significant contribution to reducing cervical cancer mortality, the analysis of Whole Slide Images (WSI), which often contain billions of pixels, remains a time-consuming and error-prone task for pathologists \cite{Bengtsson2014}. Therefore, there is an urgent need for automated cervical cell analysis methods to assist cytologists in evaluating cervical pathology images more efficiently, accurately, and objectively.

Recently, various deep learning-based detection methods have been applied in TCT screening to identify cervical abnormal cells. Specifically, Shi et al. \cite{CervicalCellClassification} developed a new classification approach for cervical cells using Graph Convolutional Networks (GCN), which aims to enhance classification accuracy by exploring the latent relationships within cervical cell images. Fei et al. \cite{RobustCervicalAbnormalCell} achieved robust detection of cervical abnormal cells through a distillation method with local-scale consistency refinement, thereby assisting pathologists in making more accurate diagnoses. Additionally, Chai et al. \cite{DPDNet} introduced an innovative dual-path discriminative detection network (DPD-Net), which effectively extracts and distinguishes features of normal and abnormal cells. These studies highlight the potential of deep learning in improving the accuracy of abnormal cervical cell detection.

Although the aforementioned methods have made significant progress in TCT screening, several issues remain to be addressed: 1) In clinical practice, pathologists typically assess target cells by comparing them with their surrounding cells, particularly when the cells exhibit ambiguous classifications. However, existing detection algorithms generally lack effective modeling of the visual feature correlations between cells, overlooking the important diagnostic information contained in spatial correlation features. 2) Cellular lesions are often accompanied by morphological abnormalities in the nucleus, such as enlargement, irregular membranes and unclear boundaries. Considering both the inter-cellular correlations and the intra-discriminative features of the cells could substantially enhance detection performance. However, no existing detection algorithm effectively combines spatial inter-correlation features with intra-discriminative features of cells due to the lack of a comprehensive fusion strategy.

To address these issues, we propose a novel mixed feature fusion network using hypergraph computation for cervical abnormal cell detection. We adopt the classic one-stage detection framework \cite{YOLO} and appropriately integrate the hypergraph computation module at multiple levels of an end-to-end detection network \cite{YOLO11}. Specifically, we use a Multi-level Fusion Sub-network (MLF-SNet) integrating three distinct types of convolution to enrich the diversity of feature representations, thereby enhancing the feature extraction ability. Then we introduce a Cross-level Feature Fusion Strategy with Hypergraph Computation module (CLFFS-HC). This strategy combines inter-correlation features of cells at the individual level with intra-discriminative features of cells at the population level, unifying them into an end-to-end framework to seamlessly integrate mixed features. Experimental results on three publicly available datasets demonstrate the superiority of our method in cervical abnormal cell detection.

\section{Method}
The proposed framework, illustrated in Fig. 1, consists of three progressive computational phases to enhance cervical abnormal cell detection. \textbf{First, in the WSI preprocessing stage}, we implement a sliding window strategy to extract 640×640 image patches from Whole Slide Images. This dimension addresses the inherent challenges of high-resolution pathological image processing while maintaining cellular-level details. \textbf{Second, for hierarchical feature extraction}, we develop a Multi-level Fusion Sub-network (MLF-SNet) to overcome the limitations of conventional one-stage detectors. Compared to standard YOLO architectures with limited multi-level feature integration capabilities, our MLF-SNet employs parallel convolution pathways to enrich the diversity of feature representations. \textbf{Third, in the feature enhancement phase}, we introduce a novel hypergraph computation approach\cite{aaai2019} through a Cross-Level Feature Fusion Strategy (CLFFS-HC). This module constructs dynamic hyperedges connecting cross-level feature representations from different network layers, combining inter-correlation features of cells at the individual level with intra-discriminative features of cells at the population level.

\begin{figure}
\includegraphics[width=\textwidth]{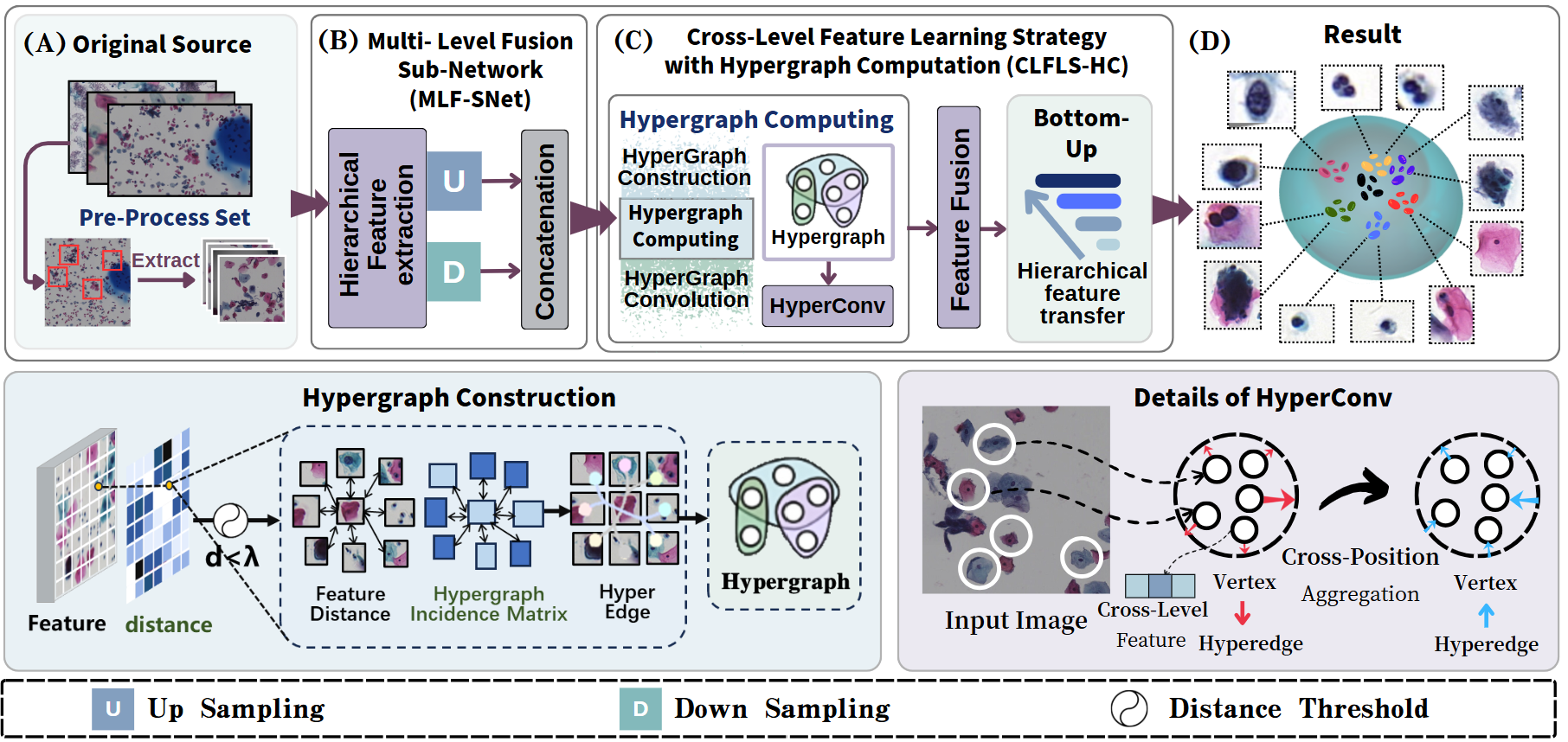}
\caption{The overview of our proposed framework,(A):Data Preprocessin, (B): MLF-SNet performs feature extraction on the preprocessed images, (C): CLFFS-HC achieves complex interactions of different types and levels of mixed features across multiple levels, resulting in high-precision detection results. The implementation of Hypergraph Construction and Hypergraph Convolution is explained in detail.} \label{fig1}
\end{figure}

\subsection{Multi-level Fusion Sub-network (MLF-SNet)}
To enhance the capture of spatial correlations between cellular structures, we introduce hypergraph computation into our design. A hypergraph $\mathcal{G}=(\mathcal{V}, \mathcal{E})$ consists of vertices $\mathcal{V}$ and hyperedges $\mathcal{E}$. We extract five feature layers from the backbone to construct the vertices $\mathcal{V}$. To fuse multi-level information, inspired by multi-level convolution \cite{duochidu}, we design a multi-level fusion sub-network during feature extraction process. This subnetwork includes a $1 \times 1$ convolution for channel feature recalibration, deformable convolution (DCN) \cite{DCNV} for effective spatial correlation information, and retains the C3K2 convolutional module to achieve hierarchical feature preservation. This approach allows the model to adapt its feature extraction and fusion techniques based on varying inputs, facilitating a comprehensive modeling of multi-level features. The detailed implementation is illustrated in Fig. 2.

\begin{figure}
    \centering
    \includegraphics[width=\textwidth]{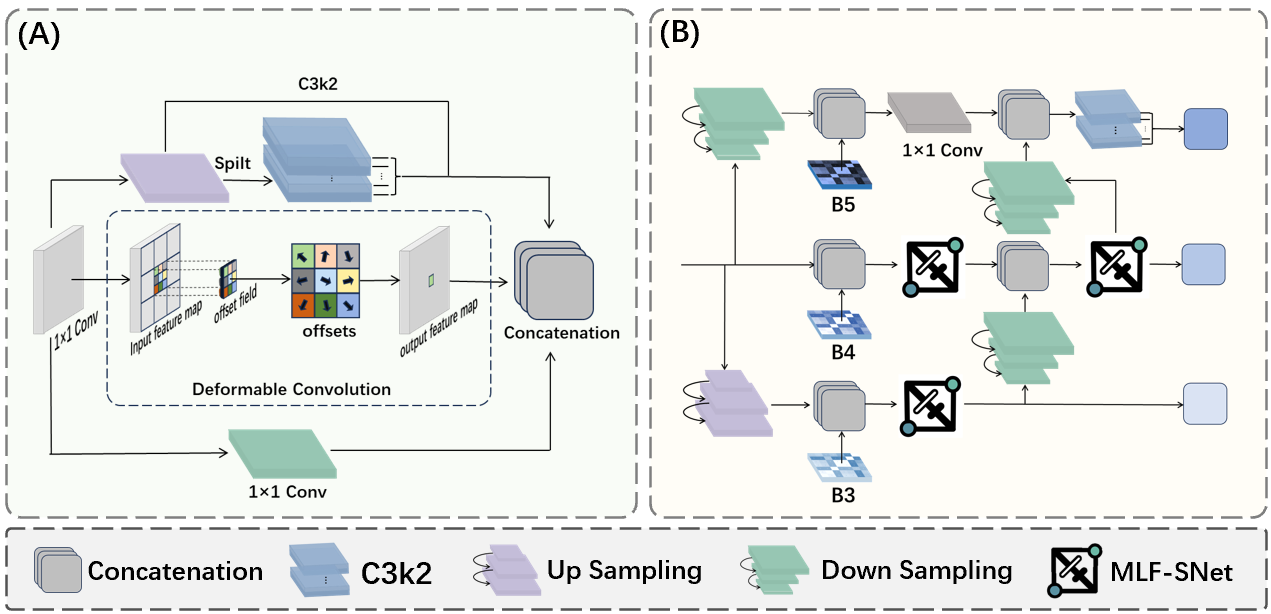}
    \caption{(A): Architecture of MLF-SNet, featuring channel recalibration (1$\times$1 conv), spatial adaptation (DCN), and hierarchical feature retention (C3K2 module); (B): Architecture of the fusion strategy and the Bottom-Up Pathway.}
    \label{fig2}
\end{figure}

As illustrated in Fig. 2(A), MLF-SNet processes raw cell images through iterative layer fusion, producing enhanced multi-scale features. A final 1×1 convolution compresses these features into an optimized output. This combination enables adaptive feature extraction while maintaining critical spatial relationships between cells. The process works synergistically with CLFFS-HC to improve detection accuracy.

\subsection{Cross-level Feature Fusion Strategy with Hypergraph Computation module (CLFFS-HC)}

To address inadequate integration of inter-correlation and intra-discriminative cellular features,  we design the Cross-level Feature Fusion Strategy with Hypergraph Computation module (CLFFS-HC). As described in Section 2.1, we construct the set of hypergraph vertices $\mathcal{V}$. To model the neighborhood relationships within the hypergraph module, we construct the set of hyperedges $\mathcal{E}$ through a distance threshold \(\lambda\). Specifically, for each feature point \(x_u\), we find all feature points that are less than $\lambda$ distance away and form a hyperedge with them and \(x_v\). A hyperedge \(e\) can be represented as:

\begin{equation}
e=\{u \mid ||x_{u}-x_{v}||_{2}<\lambda, u\in V\},
\end{equation}

The Euclidean Norm \(\| \cdot \|_2\) represents all such hyper-edges forming the hyper-edge set $E$. The incidence matrix $H$ of the hypergraph $\mathcal{G}=(\mathcal{V}, \mathcal{E})$ is defined as:

\begin{equation}
H_{ve}=\left \{\begin{matrix} 
1, & \text{if } {v} \in \mathcal{E} \\ 
0, & \text{if } {v} \notin \mathcal{E} 
\end{matrix} \right.
\end{equation}

In hypergraph convolutional networks, as the number of layers increases, information can gradually be lost during the propagation process. By introducing residual connections, the model can directly pass the input information to the output, effectively preventing information loss. To propagate high-order information within the hypergraph structure, we utilize spatial hypergraph convolution with residual connections. The computation process is as follows:

\begin{equation}
\left \{\begin{matrix} 
X_{e}=\frac{1}{|N_{v}(e)|}\sum _{v\in N_{v}(e)}X_{v}\Theta _{e}, \\ 
X'_{v}=X_{v}+\frac{1}{|N_{e}(v)|}\sum _{e\in N_{e}(v)}X_{e}
\end{matrix} \right.
\end{equation}

Where \( N_\mathcal{V}(e) \) and \( N_e(\mathcal{V}) \) represent the vertex neighborhood of the hyperedge \( e \) and the hyperedge neighborhood of the vertex \( v \), respectively, and \( \Theta_e \) is a trainable parameter. Given a feature matrix \( X \) of the vertices and the adjacency matrix \( H \) of the hypergraph, assuming \( D_\mathcal{V}\) and \( D_e\) are the degree matrices of the vertices and hyperedges, respectively, the hypergraph convolution can be expressed as:

\begin{equation}
{HyperConv}(X, H) = X + D_{v}^{-1} H D_{e}^{-1} H^{T} X \Theta
\end{equation}

 $D_{\mathcal{V}}^{-1}HD_{e}^{-1}$ calculates the normalized adjacency matrix,  $D_{\mathcal{V}}^{-1}HD_{e}^{-1}H^{T}X$ represents the aggregation of vertex features through hyperedges to capture higher-order relationships between vertices. $\Theta$ is a learnable parameter matrix used for transforming the aggregated features, enhancing the model's expressiveness. A residual connection of $X$ is also introduced to preserve the original feature information and prevent information loss.

The feature maps of the five layers extracted by the backbone are set as $B_{1}, B_{2}, B_{3}, B_{4}, B_{5}$, respectively. These features are fused through channel concatenation to obtain a mixed feature $X_{m}$ that contains information from all layers. This $X_{m}$ is then used as input for hypergraph convolution, resulting in a high-order feature representation $X_{hyper}$:

\begin{equation}
X_{hyper} = {HyperConv}(X_{m}, H)
\end{equation}

The hypergraph convolution output $X_{hyper}$ contains high-order correlation information across levels. To optimize feature integration and ensure channel alignment, we implement a fusion strategy incorporating the Bottom-Up Pathway \cite{BUP}. As shown in Fig. 2(B), $X_{hyper}$ undergoes progressive combination with the backbone features $B_{3}, B_{4}, B_{5}$ to generate multi-level outputs $N_{1}, N_{2}, N_{3}$. Unlike the original model limited to adjacent-layer fusion, our method can directly fuse the feature information extracted from the backbone. This design significantly mitigates connectivity disparities between features at different network depths. Deep layer features inherently lose spatial resolution due to increased abstraction. Our Bottom-Up Pathway addresses this limitation by strategically transferring high-resolution details from shallow layers to deeper ones, preserving critical structural information.

In summary, CLFFS-HC overcomes the limitations of traditional grid structures, enhancing the comprehensiveness and granularity of feature representations. By combining inter-correlation features of cells at the individual level with intra-discriminative features of cells at the population level, this strategy successfully integrates mixed features.

\section{Experimental Results}

\subsection{Dataset and Experimental Setup}

{\textbf{Dataset:}} To evaluate the effectiveness of the proposed model, we conducted extensive experiments on public datasets that cover different staining and imaging conditions. The data used for testing include: (A) Alibaba Tianchi competition \cite{dataset1}, (B) Comparison Detector \cite{dataset2}, and (C) CRIC \cite{dataset3}. Dataset A consists of slide images of cervical smears labeled by professional doctors, with each slide image obtained using a 20x digital scanner. Dataset B comprises 7,410 cervical cytology images with 11 types of abnormal cell lesions. Note that, for comparison with the SOTA model \cite{NN}, we only selected four types of abnormal cells for the experiment, namely ASC-US, ASC-H, LSIL, and HSIL cells. There are 400 pap smear cervical images in the dataset C in total. The nuclei of epithelial cells are addressed in six classes according to the Bethesda system, including Negative, ASC-US, ASC-H, LSIL, HSIL, and SCC.

{\textbf{Experimental Details:}} Our implementation is based on MMdetection\cite{MMD}, with all parameters optimized using Adam \cite{Adam}. The model is implemented in PyTorch on a single Nvidia Tesla P100 GPU. Performance is evaluated using two metrics: COCO-style Average Precision(AP) and Average Recall(AR). We calculate the AP over multiple IoU thresholds from 0.5 to 0.95 with a step size of 0.05, and separately evaluate the AP at an IoU threshold of 0.5 (denoted as AP.5).

\subsection{Evaluation of Cervical Abnormal Cell Detection}

{\textbf{Comparative analysis:}} We compare the performance of our proposed method with existing methods for cervical lesion detection and representative methods for object detection. To ensure fair comparison, we employed consistent preprocessing methods across all experiments for each dataset. Specifically, the sliding window method is used to crop dataset A into 640×640 images, no preprocessing is applied to dataset B, and the images of dataset C are augmented using imgaug. We will release the processed images in our publicly available code repository. The results are shown in Table 1. Our method achieves superior performance across all evaluation metrics, especially with a significant improvement in the detection performance of AP.5. These results demonstrate the effectiveness of incorporating hypergraph computation into cervical lesion detection.

\begin{table}[h]
\centering
\begin{tabular}{l@{\hspace{10pt}} c@{\hspace{10pt}} c@{\hspace{10pt}} c  @{\hspace{12pt}} c@{\hspace{10pt}} c@{\hspace{10pt}} c  @{\hspace{12pt}} c@{\hspace{10pt}} c@{\hspace{10pt}} c}
\hline
Method & \multicolumn{3}{c}{dataset A\cite{dataset1}} & \multicolumn{3}{c}{dataset B\cite{dataset2}} & \multicolumn{3}{c}{dataset C\cite{dataset3}} \\ \hline
 & AP & AP.5 & AR & AP & AP.5 & AR & AP & AP.5 & AR \\ \hline
Faster R-CNN\cite{RCNN}&  50.5&  74.1&  66.4& 18.3 & 36.6 & 33.2 &  33.6&  58.7&  52.5\\
RetinaNet\cite{Rnet} &  50.6&  80.7&  68.2& 14.9 & 29.3 & 33.8 &  32.2&  55.3&  49.5\\
Sparse R-CNN\cite{SRCNN} &  69.2&  86.5&  76.2& 18.0 & 35.0 & 44.3 &  30.6&  53.8&  52.8\\
Cascade R-CNN\cite{CRCNN} &  44.4&  69.5&  62.5& 18.4 & 34.7 & 33.0 &  35.9&  59.0&  54.7\\
YOLOX\cite{YOLOX} & 28.5 & 55.2 & 60.6 & 17.1 & 33.7 & 28.1 & 29.1 & 48.3 & 44.2 \\
RT-DETR\cite{RT} & 72.2 & 88.9 & 80.1 & 21.3 & 36.9 & 40.2 & 32.9 & 55.9 & 50.4\\
\hline
SOTA for A\cite{RT} & 72.2 & 88.9 & 80.1 &  &  &  &  &  &  \\
SOTA for B\cite{NN} &  &  &  & 24.6 & 44.7 & 46.6 &  &  &  \\
SOTA for C\cite{DPDNet}&  &  &  &  &  &  & 35.2 & 61.9 & 55.3\\
\hline
ours & 79.2& 97.5 & 91.4 & 29.8 & 51.4 & 54.4 & 44.6 & 77.0 & 72.4\\
\hline
\end{tabular}
\caption{Comparison of different methods on three datasets}
\end{table}

{\textbf{Ablation Study:}} Ablation studies were conducted on three public datasets to evaluate the contribution of each component. As shown in Table 2, our method demonstrates consistent improvements across various metrics compared to the baseline YOLO11n model. In particular, the significant improvement in recall indicates that our model has effectively addressed the issue of false negatives by incorporating the idea of hypergraph computation.

\begin{table}[h]
\centering
\begin{tabular}{l@{\hspace{10pt}} c@{\hspace{10pt}} c@{\hspace{10pt}} c  @{\hspace{12pt}} c@{\hspace{10pt}} c@{\hspace{10pt}} c  @{\hspace{12pt}} c@{\hspace{10pt}} c@{\hspace{10pt}} c}
\hline
Method & \multicolumn{3}{c}{dataset A\cite{dataset1}} & \multicolumn{3}{c}{dataset B\cite{dataset2}} & \multicolumn{3}{c}{dataset C\cite{dataset3}} \\ \hline
 & AP & AP.5 & AR & AP & AP.5 & AR & AP & AP.5 & AR \\ \hline
baseline\cite{YOLO11} & 74.0 & 86.5 & 78.4 & 21.7 & 43.0 & 47.8 & 32.6 & 56.7 & 50.7 \\
\hline
+MLF-SNet & 79.6 & 90.2 & 84.2 & 23.9 & 45.2 & 50.1 & 36.1 & 61.0 & 55.4 \\
\hline
+CLFFS-HC & 82.4 & 91.4 & 82.7 & 25.4 & 46.7 & 52.1 & 38.9 & 69.3 & 63.2 \\
\hline
+MLF-SNet\\
and CLFFS-HC & 79.2 & 97.5 & 91.4 & 29.8 & 51.4 & 54.4 & 44.6 & 77.0 & 72.4 \\
\hline
\end{tabular}
\caption{Results of the Ablation study}
\end{table}

{\textbf{Visualization Map:}} 
To visually demonstrate clinical relevance, we present comparative detection maps of representative samples in Fig. 3. Our method successfully identifies subtle lesion patterns frequently missed by conventional approaches, while maintaining superior specificity against complex backgrounds. This dual capability of reducing both false negatives and false positives significantly enhances diagnostic reliability, particularly in challenging cases with ambiguous cellular morphology. The visual evidence aligns with quantitative metrics, substantiating our method's potential as a clinically viable decision-support tool.

\begin{figure}
\includegraphics[width=\textwidth]{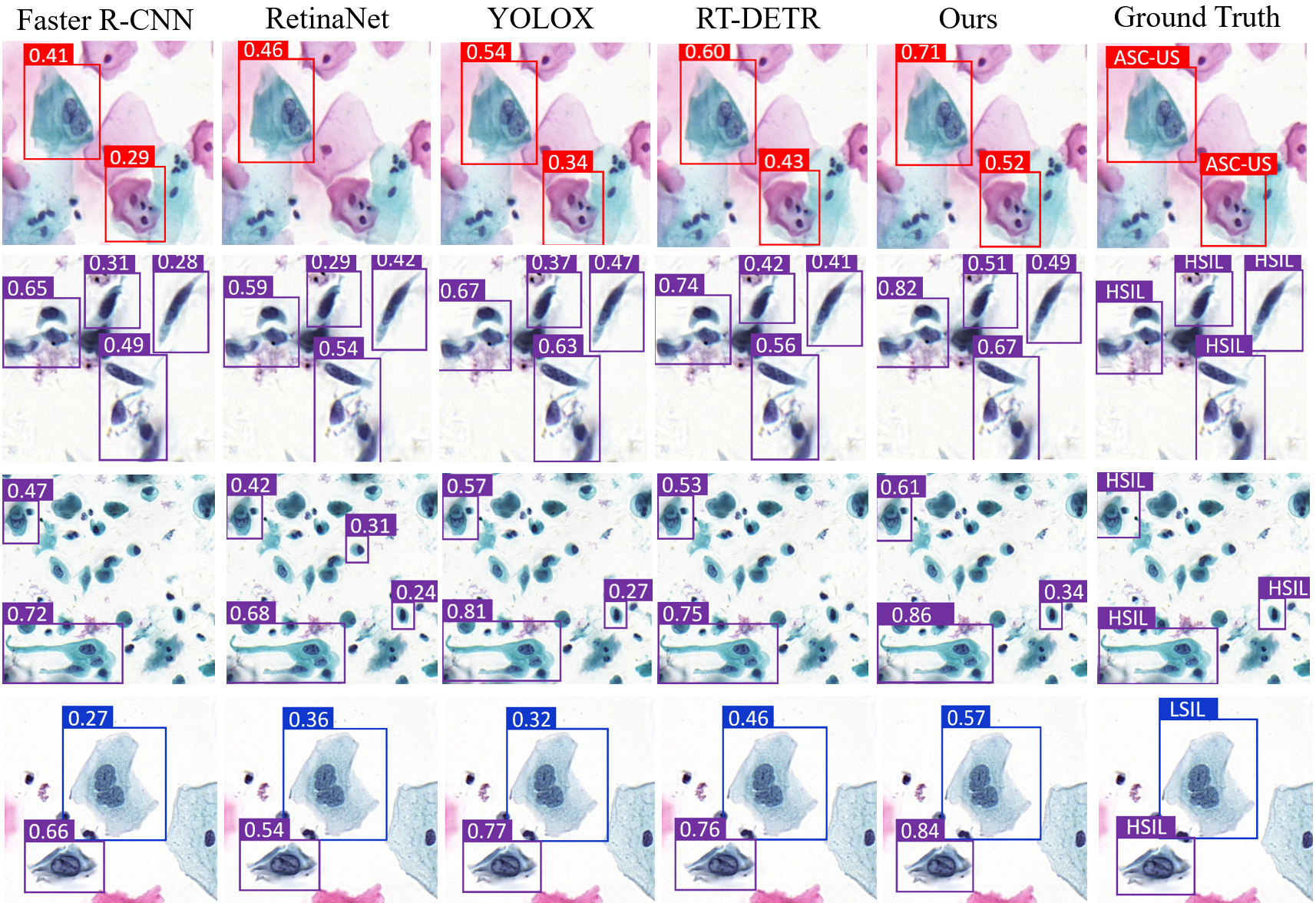}
\caption{Comparative Experiment Visualization Map} \label{fig3}
\end{figure}

\section{Conclusion}
In this paper, we have proposed a novel hypergraph-based network for cervical abnormal cell detection that effectively addresses two critical limitations in existing methods: the inadequate modeling of spatial correlations and the insufficient integration of inter-cellular and intra-cellular features. Our main contributions include: (1) the development of a Multi-level Fusion Sub-network (MLF-SNet) that enriches feature representation through diverse convolution operations, and (2) the introduction of a Cross-level Feature Fusion Strategy with Hypergraph Computation (CLFFS-HC) that effectively combines inter-correlation and intra-discriminative cellular features.
Extensive experiments on three publicly available datasets demonstrate the superior performance of our proposed method compared to existing approaches. The ablation studies further validate the effectiveness of each component in our framework, particularly highlighting the significant improvement in detection accuracy through hypergraph computation.This work provides promising directions for future research in automated cervical cell analysis. Future work could focus on extending the framework to handle more diverse pathological conditions and optimizing the computational efficiency for real-time clinical applications.

\subsubsection{\ackname} Acknowledgments. This work was supported by the Nantong Science and Technology Program Project (JC2024055), Undergraduate Innovation Training Program of Nantong University (X202510304412), Undergraduate Innovation Training Program of School of Artificial Intelligence and Computer Science(202504).

\subsubsection{\discintname} The authors have no competing interests to declare that are relevant to the content of this article.

%
\bibliographystyle{splncs04}  
\bibliography{Paper-0793}    

\end{document}